\documentclass[runningheads]{llncs}
\usepackage{graphicx}
\usepackage[caption = false]{subfig}

\usepackage{tikz}
\usepackage{comment}
\usepackage{amsmath,amssymb} 
\usepackage{color}

\usepackage{dsfont}
\usepackage{multirow}
\usepackage{cite}

\usepackage{verbatim}
\usepackage{pseudocode}

\usepackage{array}
\usepackage{booktabs, caption, makecell}

\usepackage[pagebackref=true,breaklinks=true,letterpaper=true,colorlinks,bookmarks=false]{hyperref}

\newcolumntype{L}[1]{>{\raggedright\arraybackslash}m{#1}}
\newcolumntype{C}[1]{>{\centering\arraybackslash}m{#1}}
\newcolumntype{R}[1]{>{\raggedleft\arraybackslash}m{#1}}
\newcolumntype{+}{>{\global\let\currentrowstyle\relax}}
\newcolumntype{^}{>{\currentrowstyle}}

\newcommand{\RomNum}[1]{\MakeUppercase{\romannumeral #1}}

\newcommand{\etal}{\textit{et al.}}
\newcommand{\eg}{\textit{e.g.}}

\begin{document}
\pagestyle{headings}
\mainmatter
\def\ECCVSubNumber{3397}  

\title{Semantic Line Detection Using Mirror Attention and Comparative Ranking and Matching} 

\titlerunning{Semantic Line Detection}
%
\author{Dongkwon Jin\orcidID{0000-0002-6748-3284} \and
Jun-Tae Lee\orcidID{0000-0003-2953-8851} \and \
Chang-Su Kim\orcidID{0000-0002-4276-1831}}
\authorrunning{D. Jin \etal}
%
\institute{School of Electrical Engineering, Korea University, Seoul, Korea\\
\email{\{dongkwonjin,jtlee\}@mcl.korea.ac.kr, changsukim@korea.ac.kr}}
\maketitle

\begin{abstract}
A novel algorithm to detect semantic lines is proposed in this paper. We develop three networks: detection network with mirror attention (D-Net) and comparative ranking and matching networks (R-Net and M-Net). D-Net extracts semantic lines by exploiting rich contextual information. To this end, we design the mirror attention module. Then, through pairwise comparisons of extracted semantic lines, we iteratively select the most semantic line and remove redundant ones overlapping with the selected one. For the pairwise comparisons, we develop R-Net and M-Net in the Siamese architecture. Experiments demonstrate that the proposed algorithm outperforms the conventional semantic line detector significantly. Moreover, we apply the proposed algorithm to detect two important kinds of semantic lines successfully:  dominant parallel lines and reflection symmetry axes.
Our codes are available at \href{https://github.com/dongkwonjin/Semantic-Line-DRM}{https://github.com/dongkwonjin/Semantic-Line-DRM}.

\end{abstract}

\begin{figure*}[b]

    \centering
    \subfloat {}\\[-3.5ex]
    \subfloat {\includegraphics[width=2.35cm,height=1.7cm]{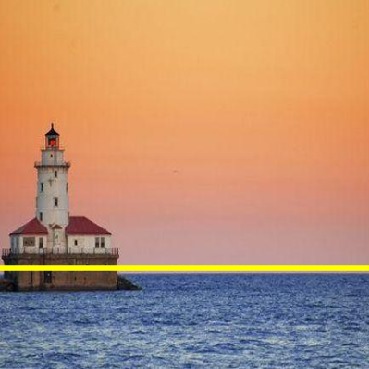}}\,\!
    \subfloat {\includegraphics[width=2.35cm,height=1.7cm]{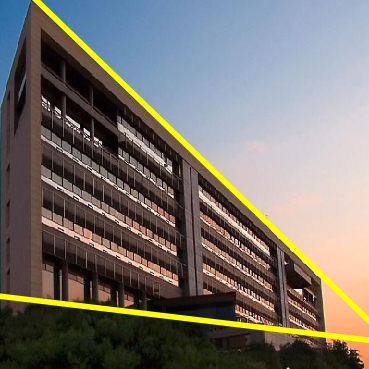}}\,\!
    \subfloat {\includegraphics[width=2.35cm,height=1.7cm]{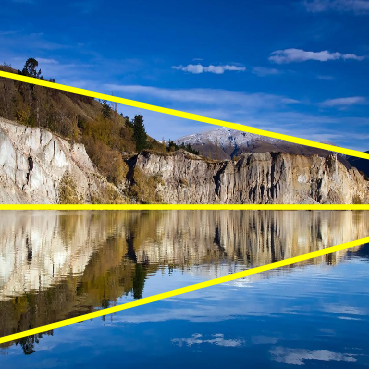}}\,\!
    \subfloat {\includegraphics[width=2.35cm,height=1.7cm]{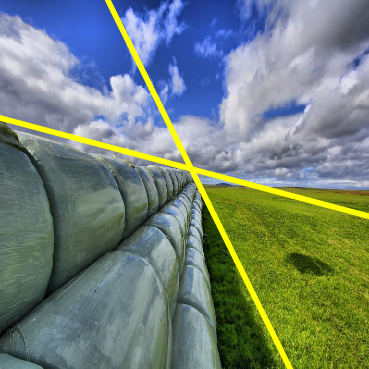}}\,\!
    \subfloat {\includegraphics[width=2.35cm,height=1.7cm]{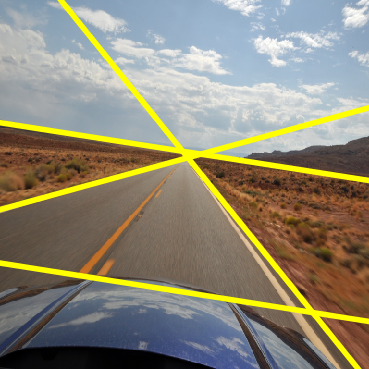}}
    \caption
    {
       Examples of various semantic lines.
    }
    \label{fig:intro}
\end{figure*}

\section{Introduction}
A \textit{semantic line}~\cite{lee2017} can be roughly defined as a dominant line, separating different semantic regions in a scene, which is reasonably approximated by an end-to-end straight line, as exemplified in Fig.~\ref{fig:intro}.

Semantic lines are essential components in high-level image understanding~\cite{Freeman2007,lee2018photographic,guo2012,hillel2014,zhou2017,zhou2019_nips}. In photography, photographic composition rules, such as horizontal, diagonal, and symmetric ones, are described by semantic lines~\cite{Freeman2007,lee2018photographic}. Under perspective projection, dominant parallel lines in the 3D world are projected to semantic lines in 2D images, intersecting at vanishing points and conveying depth impressions~\cite{zhou2017}. Also, in autonomous driving systems~\cite{guo2012,hillel2014}, the boundaries of road lanes, sidewalks, or crosswalks are important sematic lines. However, it is difficult to detect semantic lines, which are often unobvious and implied by complex boundaries of semantic regions.

Although many techniques have been developed to detect lines by exploiting low-level cues~\cite{matas2000,von2008,desolneux200,akinlar2011} or deep features~\cite{huang2018,xue2019,zhou2019_line}, they extract many short (possibly noisy) line segments or rather obvious lines in man-made environments. Also, several attempts~\cite{workman2016,zhai2016,koo2013} have been made to detect unobvious horizon lines. However, horizons are just a specific type of semantic lines.  Recently, SLNet, which is a general semantic line detector, was proposed in~\cite{lee2017}. Although SLNet provides promising results, it tends to detect many redundant lines near the boundaries of semantic regions.

\begin{figure*}[t]

  \centering
  \includegraphics[width=1\linewidth]{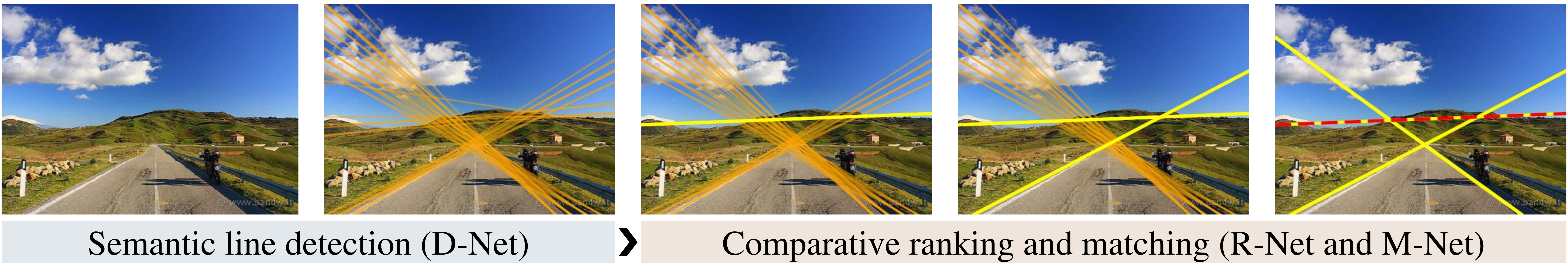}\\[-5pt]

  \caption{Illustration of the proposed DRM algorithm: First, D-Net extracts semantic lines (orange). Second, R-Net selects the most semantic line (yellow). Third, M-Net removes redundant lines overlapping with the selected one. The second and third steps are iteratively applied. In this example, three semantic lines are selected. The first one is called the primary semantic line (dashed red).}
  \label{fig:semantic_line_fig}

\end{figure*}

In this paper, we propose a novel semantic line detection algorithm, called DRM, which consists of three networks: detection network with mirror attention (D-Net) and comparative ranking and matching networks (R-Net and M-Net). In Fig.~\ref{fig:semantic_line_fig}, D-Net first extracts semantic lines by classifying and regressing candidate lines. For effective detection, we design the mirror attention module and the region pooling layer in D-Net. Then, by comparing semantic lines in pairs, R-Net selects the most semantic line and M-Net removes redundant lines overlapping with the selected one. This comparative ranking and matching process is performed iteratively. In Fig.~\ref{fig:semantic_line_fig}, three iterations are performed to yield three semantic lines. Experimental results demonstrate that the proposed DRM algorithm outperforms the conventional SLNet~\cite{lee2017} significantly.

This work has the following major contributions:

\begin{itemize}
\item We develop D-Net to detect semantic lines, in which the mirror attention module and the region pooling layer extract discriminate features effectively.
\item We propose two Siamese networks, R-Net and M-Net, for pairwise ranking and matching of semantic lines.
\item We construct a challenging dataset (SEL\_Hard) of semantic lines, which are highly implied in cluttered scenes.\footnote{SEL\_Hard is available at \href{https://github.com/dongkwonjin/Semantic-Line-DRM}{https://github.com/dongkwonjin/Semantic-Line-DRM}.}
\item We also apply the proposed algorithm to two important line detection tasks: dominant parallel lines and reflection symmetry axes.
\end{itemize}

\section{Related Work}

\subsection{Line detection}
Lines are geometrically important cues to describe the layouts or structural information of images. In line segment detection \cite{matas2000,von2008,desolneux200,akinlar2011}, many short line segments are detected using low-level cues (\eg~image gradients). However, this approach  may not discriminate meaningful lines from noisy ones. To utilize higher-level cues, deep learning methods have been proposed \cite{huang2018,xue2019,zhou2019_line,sun2019}.
In \cite{huang2018}, two networks were used to predict, respectively, a line heat map and junctions in man-made environments. Then, the wireframe, summarizing the scene, was obtained by connecting the junctions based on the heat map. In \cite{zhou2019_line}, a network verified whether a candidate line was salient or not, where the candidate was also generated by connecting two junctions. In \cite{xue2019}, the line segment detection was posed as the dual problem of region coloring to address local ambiguity and class imbalance. In \cite{sun2019}, a network was trained to yield the coordinates of a bounding box, whose diagonal was the resultant line segment. However, these methods \cite{huang2018,xue2019,zhou2019_line,sun2019} detect rather obvious lines in man-made environments.

Meanwhile, several methods~\cite{workman2016,zhai2016,lee2017,diaz2019} have been developed to detect implied lines. In~\cite{workman2016}, horizon lines were directly estimated by CNNs, without requiring geometric constraints. In~\cite{zhai2016}, horizons were detected similarly to~\cite{workman2016}, but their locations were refined by exploiting vanishing points. In~\cite{diaz2019}, soft labels of horizon line parameters were used to train the regression network. In~\cite{lee2017}, the first semantic line detector was proposed, which can detect general, semantically meaningful lines. Semantic lines, located near the boundaries of semantic regions, represent the layout and composition of an image, even when the boundaries are not straight lines.

\begin{figure*}[t]

  \centering
  \includegraphics[width=1\linewidth]{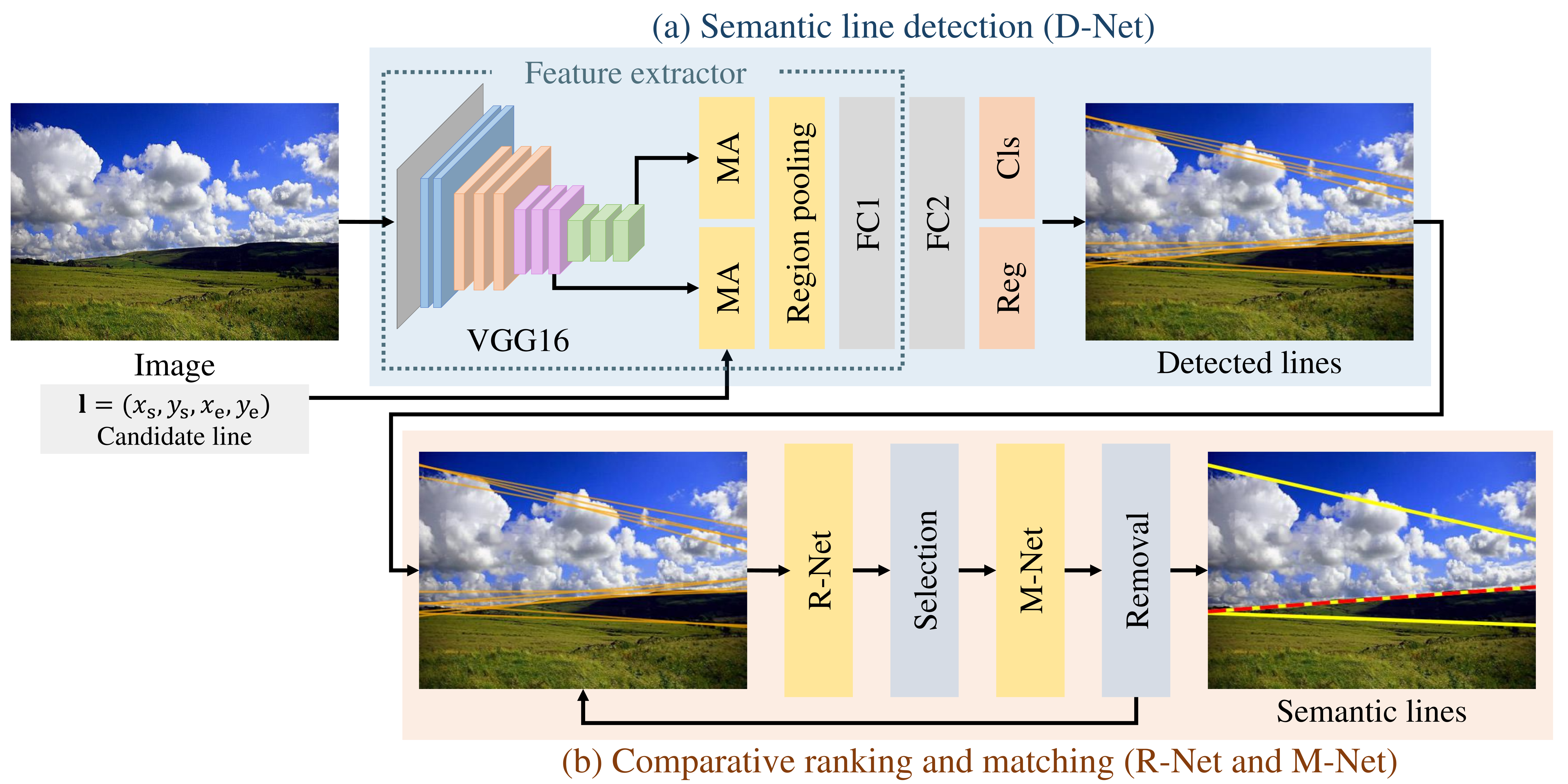}
  \caption{An overview of the proposed algorithm: (a) D-Net detects semantic lines, by classifying and regressing candidate lines, based on mirror attention (MA). (b) R-Net selects the most meaningful semantic line and M-Net removes redundant lines alternately through pairwise comparisons.}
  \label{fig:overview_fig}

\end{figure*}

\subsection{Attention mechanisms in CNNs}
Human visual system pays more attention to salient parts of a scene for efficiency \cite{corbetta2002control,itti1998model}. Similarly, attempts have been made to bias the processing resource of a neural network towards more informative parts of input data \cite{bahdanau2015neural,Ashish2017,zambaldi2018deep,yu2018qanet}. Also, attention mechanisms have been developed to improve the representation power of convolutional layers in CNN-based vision tasks \cite{wang2017,hu2018,woo2018,Park2018_BMVC,Zhao2019_CVPR}. In \cite{wang2017}, Wang~\etal~adopted an encoder-decoder structure to obtain a pixel-wise attention mask of a convolutional feature map. In~\cite{hu2018}, to address the interdependencies of filter responses, Hu~\etal~used the average-pooled feature at each channel to compute the channel-wise attention. In \cite{woo2018,Park2018_BMVC}, this channel-wise attention module has been modified to obtain both spatial and channel-wise attention. In \cite{Jetley2018_ICLR}, multiple attention maps were obtained from intermediate convolutional layers, and then the ensemble of those maps was applied to the last layer. In \cite{Zhao2019_CVPR}, Zhao and Wu applied spatial attention to lower layers to focus on local details and channel-wise attention to higher layers to capture contextual cues.

\subsection{Metric learning and order learning}
Metric learning~\cite{sohn2016,oh2016} constructs an appropriate feature embedding space, where similar objects are located tightly  while dissimilar objects are far from one another. In contrast, in order learning~\cite{lim2020}, embedded features are ordered according to the ranks or priorities of  objects. Both the similarity and order relationships depend on target applications and are implicitly defined by user-provided examples. Accordingly, the learned metric is useful for matching similar objects, \eg, in image retrieval~\cite{hoi2010,gao2014soml}, person re-identification~\cite{ding2015,chu2019}, and few-shot learning~\cite{vinyals2016_nips,sung2018_CVPR}. On the other hand, the learned order can be used to rank or sort objects, as done in image quality assessment~\cite{kong2016,Liu2017_ICCV}, object detection~\cite{tan2019,singh2019}, and age estimation~\cite{Chen2017_CVPR,chen2017_tmm}.

\section{Proposed Algorithm}
We propose a novel semantic line detection algorithm, called DRM, which is composed of three networks: D-Net, R-Net, and M-Net. Fig.~\ref{fig:overview_fig} is an overview of the proposed DRM algorithm. First, we generate candidate lines by connecting two points, uniformly sampled on image boundaries~\cite{lee2017}. Second, D-Net extracts semantic lines by classifying and regressing the candidate lines. For discriminative feature extraction, we design the mirror attention module and the region pooling layer. Third, through pairwise comparisons, we iteratively select the most meaningful semantic line and remove the other semantic lines overlapping with the selected one. For this purpose, we develop R-Net and M-Net in the Siamese architecture.

\subsection{D-Net: Semantic line detection with mirror attention}

\begin{figure*}[t]

    \centering
    \subfloat{\includegraphics[width=1.98cm,height=1.5cm]{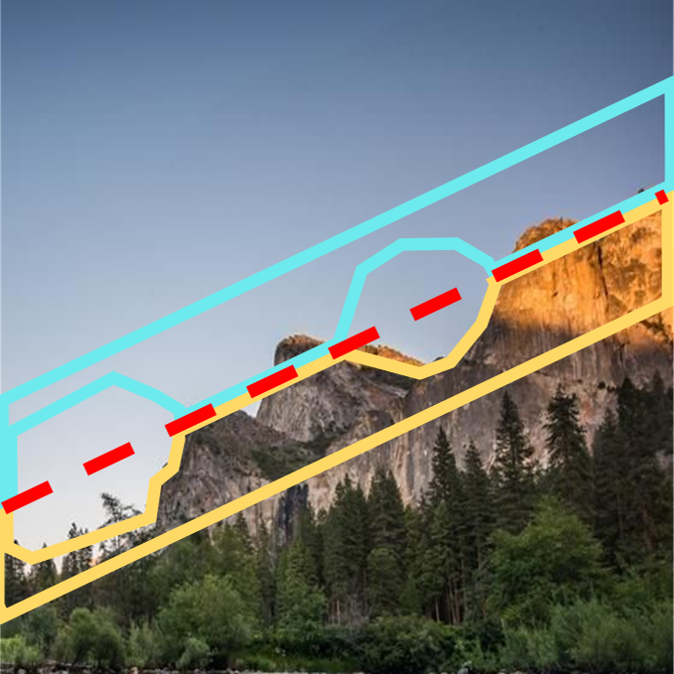}}\,\!\!
    \subfloat{\includegraphics[width=1.98cm,height=1.5cm]{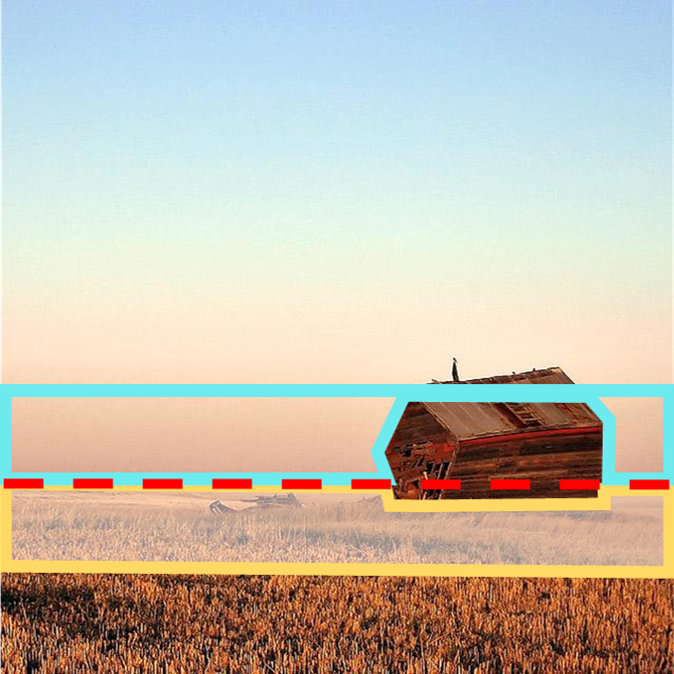}}\,\!\!
    \subfloat{\includegraphics[width=1.98cm,height=1.5cm]{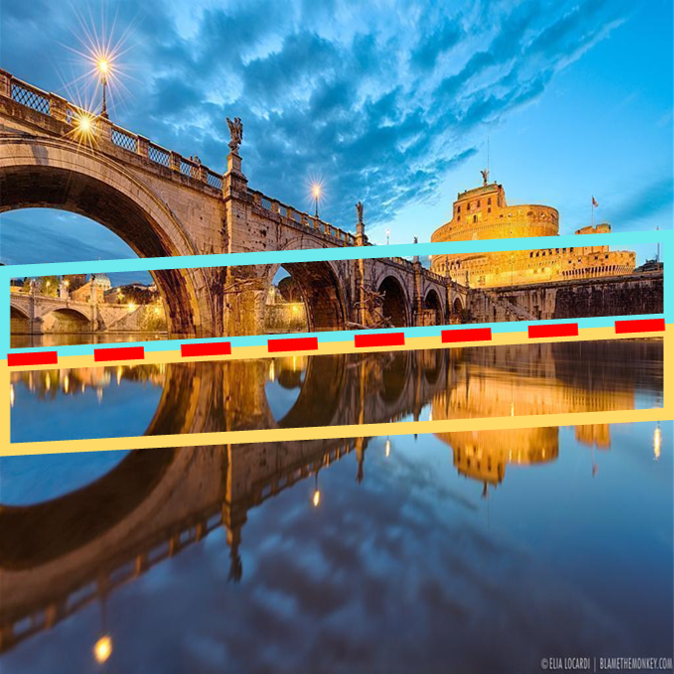}}\,\!\!    \subfloat{\includegraphics[width=1.98cm,height=1.5cm]{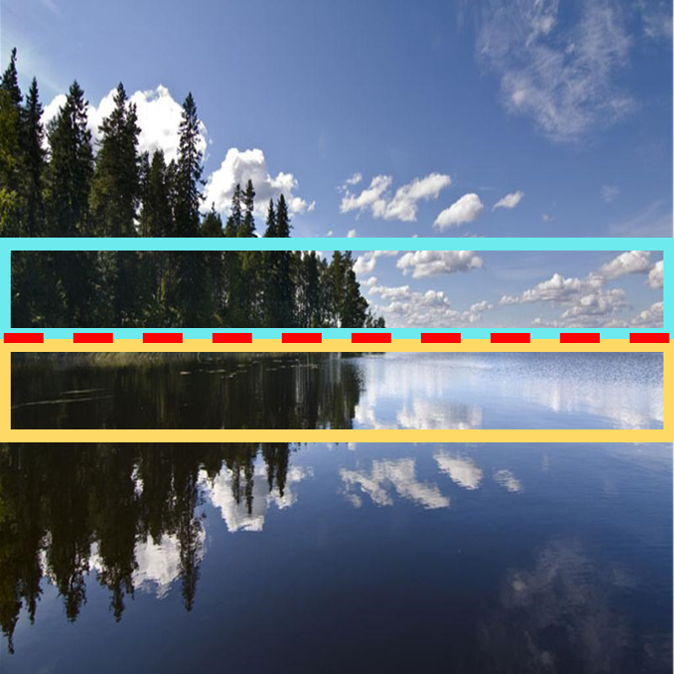}}\,\!\!
    \subfloat{\includegraphics[width=1.98cm,height=1.5cm]{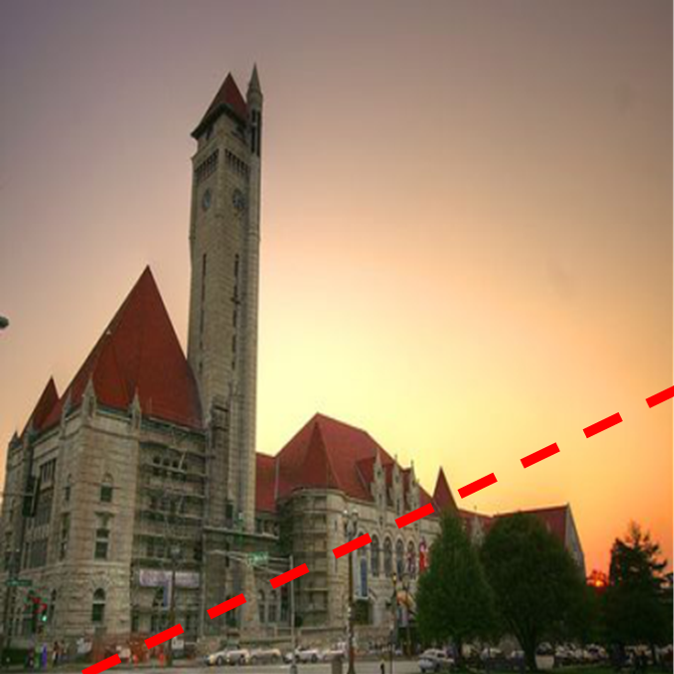}}\,\!\!
    \subfloat{\includegraphics[width=1.98cm,height=1.5cm]{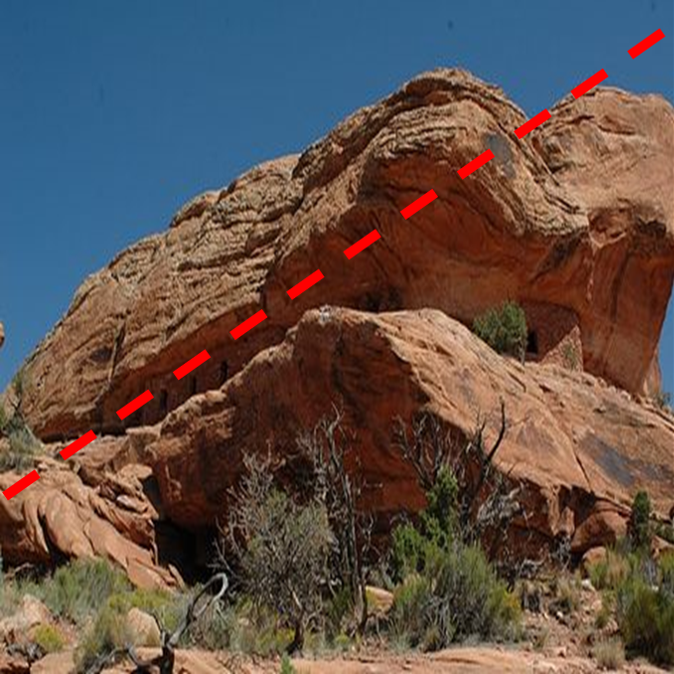}}\\[-2.5ex]
    \addtocounter{subfigure}{-6}
    \subfloat[]{\includegraphics[width=1.98cm,height=1.5cm]{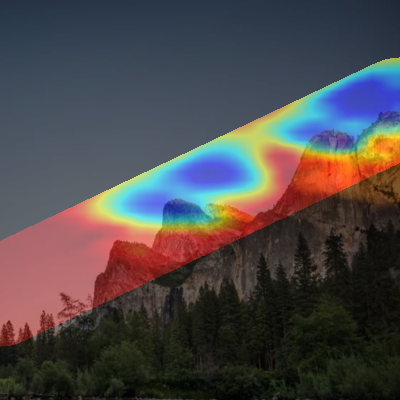}}\,\!\!
    \subfloat[] {\includegraphics[width=1.98cm,height=1.5cm]{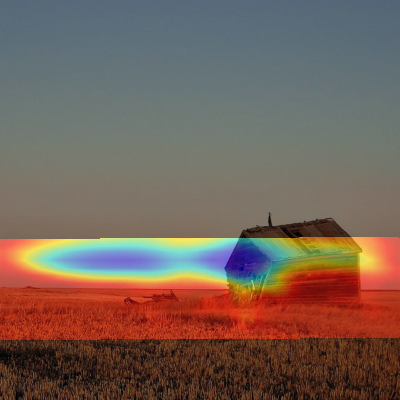}}\,\!\!
    \subfloat[] {\includegraphics[width=1.98cm,height=1.5cm]{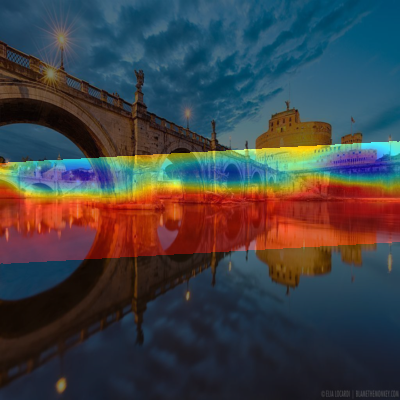}}\,\!\!
    \subfloat[] {\includegraphics[width=1.98cm,height=1.5cm]{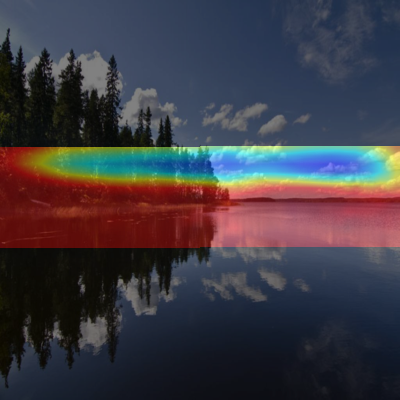}}\,\!\!
    \subfloat[] {\includegraphics[width=1.98cm,height=1.5cm]{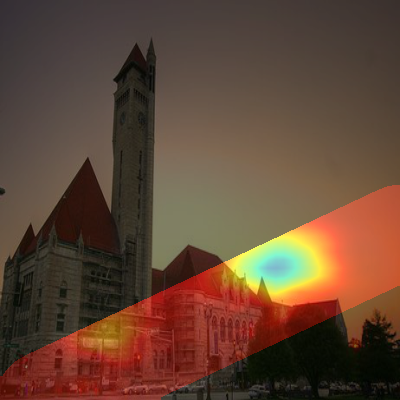}}\,\!\!
    \subfloat[] {\includegraphics[width=1.98cm,height=1.5cm]{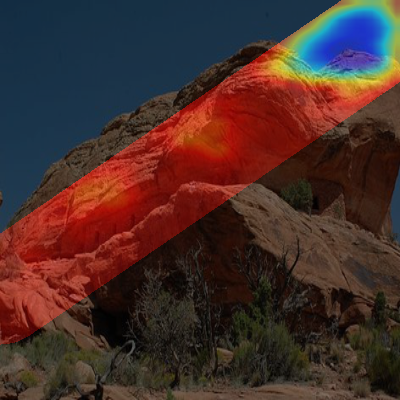}}\\[-1ex]
    \caption
    {
      [Top] A semantic (or candidate) line is shown in red, while two regions producing the line are in cyan and yellow. [Bottom] The attention mask is color-coded: red and blue depict big and small values. Note that the two regions are semantically different from each other in (a) and (b) and symmetric in (c) and (d). In (e) and (f), the candidate lines are not semantic.
    }
    \label{fig:Attmap}
\end{figure*}

\begin{figure*}[t]

  \centering
  \includegraphics[width=1\linewidth]{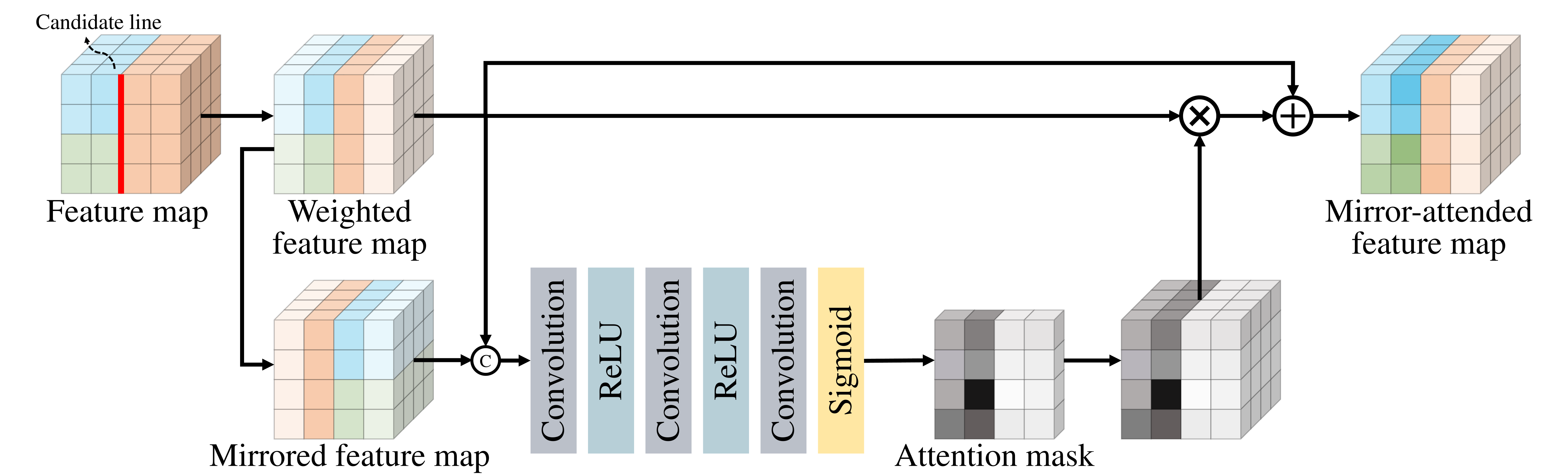}

  \caption{Illustration of the mirror attention process.}
  \label{fig:mirror_attention}

\end{figure*}

\noindent\textbf{Mirror attention:} A semantic line separates a region into two distinct sub-regions. Those two regions can be semantically different as in Fig.~\ref{fig:Attmap}(a), (b), or symmetric around the line as in Fig.~\ref{fig:Attmap}(c), (d). In the former case, if a region is mirrored along the line, it should contain quite different objects from the other region. In the latter case, it should be almost identical with the other region. To summarize, the line is semantic because of the mirrored dissimilarity (heterogeneity of two regions) or mirrored similarity (symmetry). Based on this observation, we develop the mirror attention module in Fig.~\ref{fig:mirror_attention}. Given a feature map, the mirror attention module generates an attention mask, which is then used to reweight the feature map to make it more discriminative.

We apply the mirror attention module to a convolutional feature map $X=[X^1, X^2, \ldots, X^C]\in \mathbb{R}^{H\times W\times C}$, where $H$, $W$, and $C$ denote the height, the width, and the number of channels. Since pixels near a candidate line are more relevant for semantic line detection, we first obtain a weighted feature map $Y=[Y^1, Y^2, \ldots, Y^C]\in \mathbb{R}^{H\times W\times C}$, where $X^c(k)$ for pixel $k$ is weighted by
\begin{equation}\label{eq:filter}
    Y^c({k}) = \omega(d_k)\times X^c({k}).
\end{equation}
Here, $d_k$ is the distance of pixel $k$ from the candidate line and $\omega(\cdot)$ is the Gaussian weighting function.

To analyze the mirror relationships around the candidate line, we obtain the mirrored feature map $\tilde{Y}$ by flipping $Y$ across the line. If a flipped pixel is outside the feature map, it is set to zero. Then, we concatenate $Y$ and $\tilde{Y}$ and obtain an initial mask $A_0\in \mathbb{R}^{H\times W}$ by
\begin{equation}\label{eq:channel_conv}
    A_0 = f_0([Y, \tilde{Y}])
\end{equation}
where $f_0$ is a convolutional layer using a single filter of size $n\times n\times 2C$. We set $n$ to 3 empirically. To increase the receptive field and capture the semantics from a wider region, we use two more convolution layers to yield $A_2 = f_2(f_1(A_0))$, where $f_1$ or $f_2$ uses a single filter of size $(2n+1)\times (2n+1)\times 1$. Finally, we obtain the attention mask $A=\sigma(A_2)$ where $\sigma(\cdot)$ is the sigmoid activation function.

We apply the mirror attention module to two deep layers of D-Net, as shown in Fig.~\ref{fig:overview_fig}. The attention mask may over-suppress the values in the weighted feature map $Y$. To prevent this, we adopt the residual attention scheme~\cite{wang2017}. More specifically, we obtain the attended feature map $Y_{\rm att} = [Y^1_{\rm att}, Y^2_{\rm att}, \ldots, Y^C_{\rm att}]$, where $Y^c_{\rm att}$ is given by
\begin{equation}\label{eq:residual_attention}
    Y^c_{\rm att} = (1 + A) \otimes Y^c
\end{equation}
for each $1\leq c \leq C$.

Fig.~\ref{fig:Attmap} shows examples of attention masks. In Fig.~\ref{fig:Attmap}(a)$\sim$(d), there are roughly two semantic regions around the candidate line. We see that one region is attended with small weights, while the other with big weights. Thus, the feature difference between the two regions is emphasized, facilitating the semantic line detection. Note that the mirror attention module is trained in an end-to-end manner such that emphasizing masks are generated in both cases of mirrored dissimilarity (Fig.~\ref{fig:Attmap}(a) or (b)) and mirrored similarity (Fig.~\ref{fig:Attmap}(c) or (d)). On the other hand, in Fig.~\ref{fig:Attmap}(e) or (f), a less informative mask is generated because the candidate line is not semantic.

\noindent\textbf{Region pooling:} Whereas the conventional algorithm~\cite{lee2017} uses a line pooling layer, we design a region pooling layer to extract more discriminated features from the mirror-attended feature map $Y_{\rm att}$. We set two adjacent regions ${\cal U}$ and ${\cal V}$ along the candidate line, which contain pixels whose distances from the line are less than a threshold, respectively. Then, we aggregate the regional information of ${\cal U}$ and ${\cal V}$ into $\bf u$ and $\bf v$ $\in \mathbb{R}^C$;
\begin{equation}
    {\bf u} = \frac{1}{|{\cal U}|}\sum_{{k} \in {\cal U}}{Y_{\rm att}({k})} \quad \mbox{and} \quad
    {\bf v} = \frac{1}{|{\cal V}|}\sum_{{k} \in {\cal V}}{Y_{\rm att}({k})}.
\end{equation}
Then, $\bf u$ and $\bf v$ are concatenated to form the feature vector of the candidate line.

\noindent\textbf{D-Net architecture:} We plug the proposed mirror attention module and region pooling layer into the classification-regression framework of~\cite{lee2017}. Hence, D-Net takes an image and a candidate line, parameterized by ${\bf l} = (x_s, y_s, x_e, y_e)$, and yields classification and regression results. Fig.~\ref{fig:overview_fig}(a) shows its architecture. We use the 13 convolution layers of VGG16~\cite{Simonyan2015} as the backbone, and implement two mirror attention modules after Conv10 and Conv13, respectively. From each mirror-attended feature map, the region pooling layer extracts the feature vector. The two vectors are concatenated and fed into fully connected layers FC1 and FC2. Finally, D-Net branches into two parallel output layers: one for classifying the candidate line (Cls), and the other for computing regression offsets for the line parameters (Reg). Cls computes the softmax vector ${\bf p} = (p, q)$, where $p$ is the probability that the candidate line ${\bf l}$ is semantic. Reg outputs a line offset $\Delta{\bf l}$. When $p > 0.5$, D-Net declares that the regressed line ${\bf l} + {\Delta\bf l}$ is semantic.

To train D-Net, when a candidate line is annotated by $\bar{\bf p}$ and $\Delta \bar{\bf l}$, we minimize the loss
\begin{equation}\label{loss_all}
   L({\bf p}, \bar{\bf p}, \Delta {\bf l},  \Delta \bar{\bf l}) = L_{\rm cls}({\bf p}, \bar{\bf p}) + \lambda L_{\rm reg}(\Delta {\bf l}, \Delta \bar{\bf l})
\end{equation}
where $L_{\rm cls}({\bf p}, \bar{\bf p})$ and $L_{\rm reg}(\Delta {\bf l},  \Delta \bar{\bf l})$ are the classification loss and the regression loss, respectively, and $\lambda$ is a balancing parameter. $L_{\rm cls}$ is the cross-entropy loss over the two classes (semantic and non-semantic). $L_{\rm reg}(\Delta {\bf l}, \Delta \bar{\bf l})=\eta(\Delta {\bf l} - \Delta \bar{\bf l}),$ where $\eta$ is the smooth $L_1$ loss~\cite{girshick2015fast}.

\subsection{R-Net and M-Net: Comparative ranking and matching}
\label{ssec:CRM}

Note that D-Net detects many semantic lines from densely sampled candidate lines, as illustrated in Fig.~\ref{fig:semantic_line_fig}. Since each candidate is tested independently, semantically identical lines are detected closely. For example, in Fig.~\ref{fig:semantic_line_fig}, there are three groups, each of which contains semantically identical lines. From each group, we select the most reliable line, while removing the other redundant ones. To this end, we develop the comparative ranking and matching networks, referred to as R-Net and M-Net, respectively. Given a pair of semantic lines, R-Net finds which one is more reliable and M-Net determines whether they are semantically identical or not. Thus, R-Net is related to \textit{priority} in order learning \cite{lim2020}, while M-Net is to \textit{similarity} in metric learning \cite{sohn2016,oh2016}.

\begin{figure*}[t]

  \centering
  \includegraphics[width=1\linewidth]{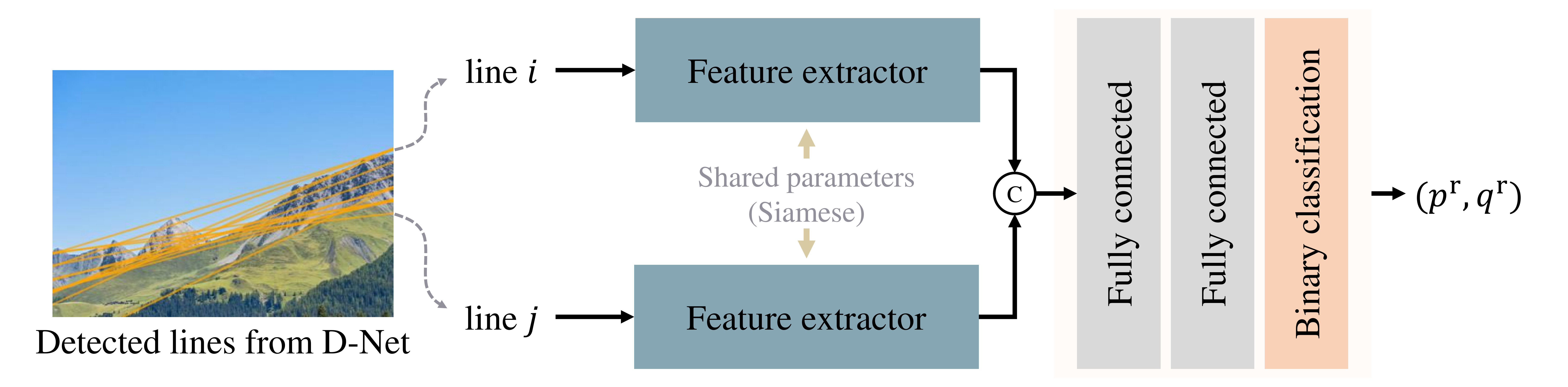}
  \caption{Siamese architecture for R-Net: Given a pair of detected lines from D-Net, R-Net decides whether one line is more reliable or less reliable than the other.}
  \label{fig:matching_fig}
\end{figure*}

Both R-Net and M-Net are implemented as binary classifiers in the Siamese architecture. Fig.~\ref{fig:matching_fig} shows R-Net, which yields a softmax probability vector ${\bf p}^{\rm r}=(p^{\rm r}, q^{\rm r})$. Here, $p^{\rm r}$ or $q^{\rm r}$ is the probability that line $i$ is more reliable or less reliable than line $j$, respectively. Since D-Net is well-trained for semantic line detection, we truncate it before the FC2 layer and use it as the feature extractor of R-Net. Then, the features of the two lines are concatenated, propagated into two fully connected layers, and categorized into one of the binary classes. The cross-entropy loss is used to train R-Net. Also, M-Net is implemented in the same way as R-Net, except that it yields a softmax vector ${\bf p}^{\rm m}=(p^{\rm m}, q^{\rm m})$, where $p^{\rm m}$ or $q^{\rm m}$ is the probability that the two lines are semantically identical or not, respectively.

Using R-Net and M-Net, we perform the selection and removal of semantic lines alternately. At step $t$, we measure the reliability of each semantic line. Specifically, the reliability $r_i$ of semantic line $i$ is defined as
\begin{equation}\label{eq:reliabilty}
    r_i = \sum_{j=1,\, j\neq i}^{N_t} p^{\rm r}_{ij}
\end{equation}
where $N_t$ is the number of available semantic lines at step $t$, and $p_{ij}^{\rm r}$ is the probability that line $i$ is more reliable than line $j$. Then, we select the most reliable line $i^*$ by
\begin{equation}\label{eq:top_rank}
    i^* = \arg \max_i r_i.
\end{equation}
We then remove the lines that are semantically identical with line $i^*$. Specifically, line $j$ is removed if the matching probability $p_{i^* j}^{\rm m}$ from M-Net is higher than 0.5. We iteratively perform the alternate selection and removal until $N_t=0$. In the example of Fig.~\ref{fig:semantic_line_fig}, three iterations are performed to select the three resultant lines. The firstly selected line is called the primary line.

We configure the training data for R-Net and M-Net as follows. Note that a detected semantic line is declared to be correct if its mean intersection over union (mIoU) ratio with the ground-truth is higher than 0.85 \cite{lee2017}. After training D-Net on the SEL dataset~\cite{lee2017}, we use the correctly detected semantic lines in the training images to train R-Net and M-Net. For R-Net, a ground-truth semantic line and one of its detection results are used as an input pair. To encode the one-hot vector $\bar{\bf p}^{\rm r}=(\bar{p}^{\rm r}, \bar{q}^{\rm r})$, the ground-truth line is used as the more reliable one than the detection result. For M-Net, we use two detected lines as input. To encode $\bar{\bf p}^{\rm m}=(\bar{p}^{\rm m}, \bar{q}^{\rm m})$, if the two lines correspond to the same ground-truth, they are regarded as semantically identical.

\section{Experimental Results}

\subsection{Datasets}

\noindent\textbf{SEL:}
The semantic line (SEL) dataset~\cite{lee2017} contains 1,750 outdoor images in total, which are split into 1,575 training and 175 testing images. Each semantic line is annotated by the coordinates of the two end-points on an image boundary. If an image has a single dominant line, it is set as the ground truth primary semantic line. If an image has multiple semantic lines, the line with the best rank by human annotators is set as the ground-truth primary line, and the others as additional ground-truth semantic lines. In SEL, 61\% of the images contain multiple semantic lines.

%

\noindent\textbf{SEL\_Hard:}
In addition to the SEL dataset, we construct a more challenging test dataset, called SEL\_Hard. Its semantic lines are more implied (or less obvious), are more severely occluded, and are in more cluttered scenes. We collect 300 images from the ADE20K image segmentation dataset~\cite{zhou2017_ade}, manually annotate semantic lines, and then also select primary lines. Notice that SEL\_Hard is constructed for testing semantic line detectors and is not used for training them. The supplemental document describes the annotation process in detail and provides example images.

\subsection{Semantic line detection results}
We assess primary and multiple semantic line detection performances on the SEL and SEL\_Hard datasets.

We measure the accuracy for primary semantic line detection and the precision and recall rates for multiple semantic line detection, based on the mIoU metric~\cite{lee2017}. A semantic line is regarded as correctly detected if its mIoU score with the ground-truth is greater than a threshold~$\tau$. Then, the accuracy of the primary semantic line detection is defined as
\begin{equation}\label{eq:acc}
    {\rm Accuracy} = \frac{N_c}{N}
\end{equation}
where $N_c$ is the number of the test images whose primary semantic lines are correctly detected, and $N$ is the number of all test images. For the multiple semantic line detection, the precision and the recall are computed by
\begin{equation}\label{eq:pre_rec}
    {\rm Precision} = \frac{N_l}{N_l + N_e}, \;\;\; {\rm Recall} = \frac{N_l}{N_l + N_m}
\end{equation}
where $N_l$ is the number of correctly detected semantic lines, $N_e$ is the number of false positives, and $N_m$ is the number of false negatives.

\begin{figure*}[t]
    \centering
    \subfloat {\includegraphics[width=4cm, height=3.5cm]{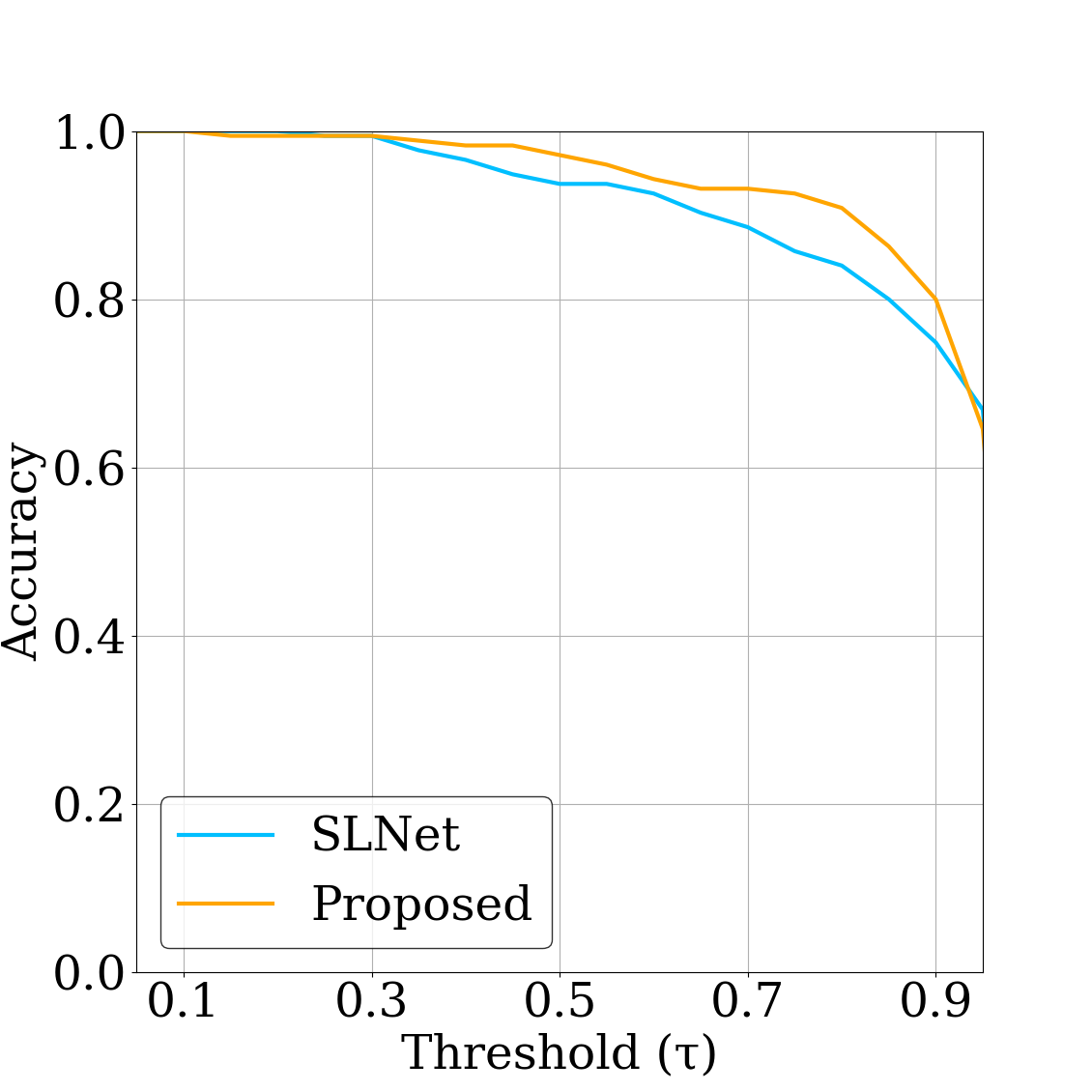}}\,\!\!
    \subfloat {\includegraphics[width=4cm, height=3.5cm]{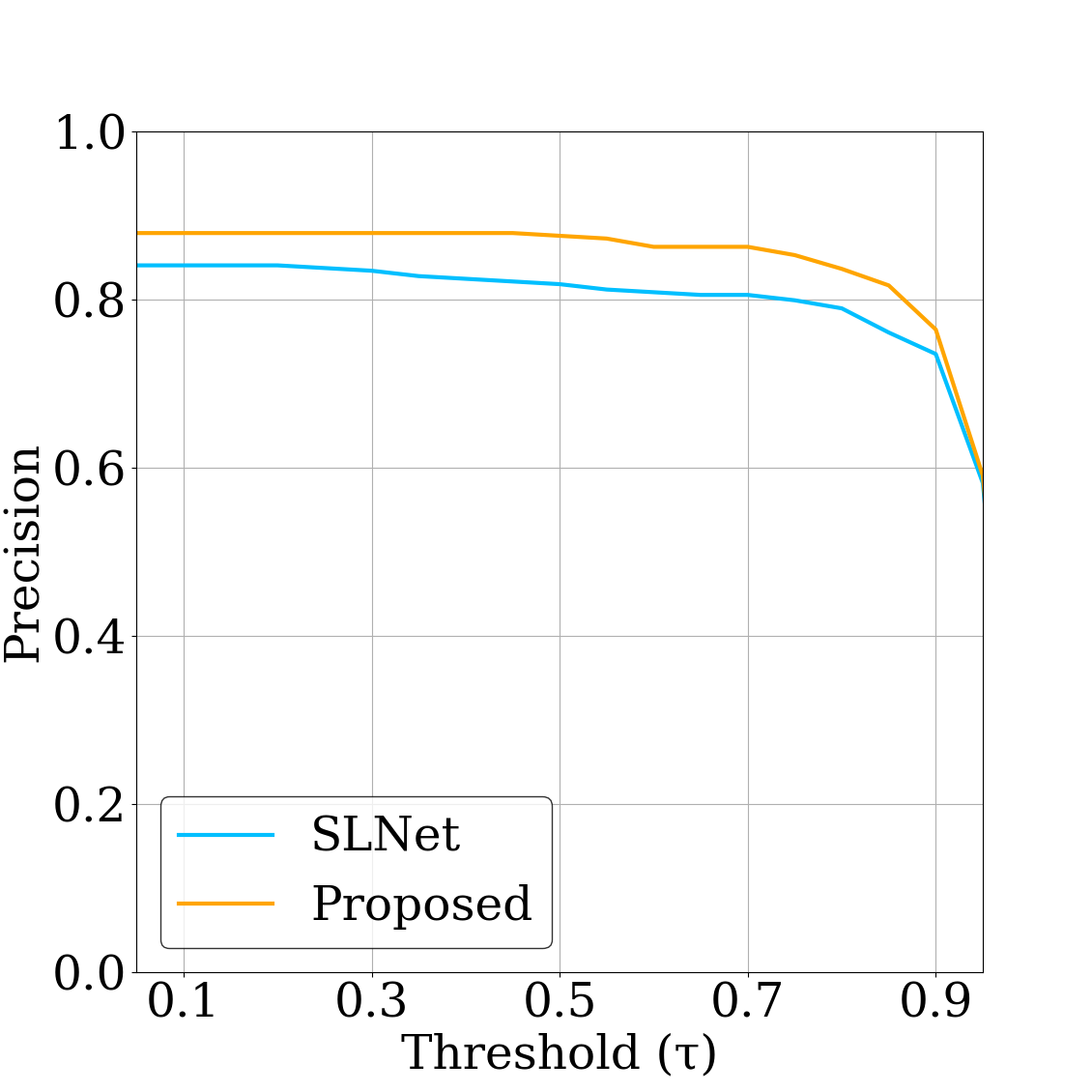}}\,\!\!
    \subfloat {\includegraphics[width=4cm, height=3.5cm]{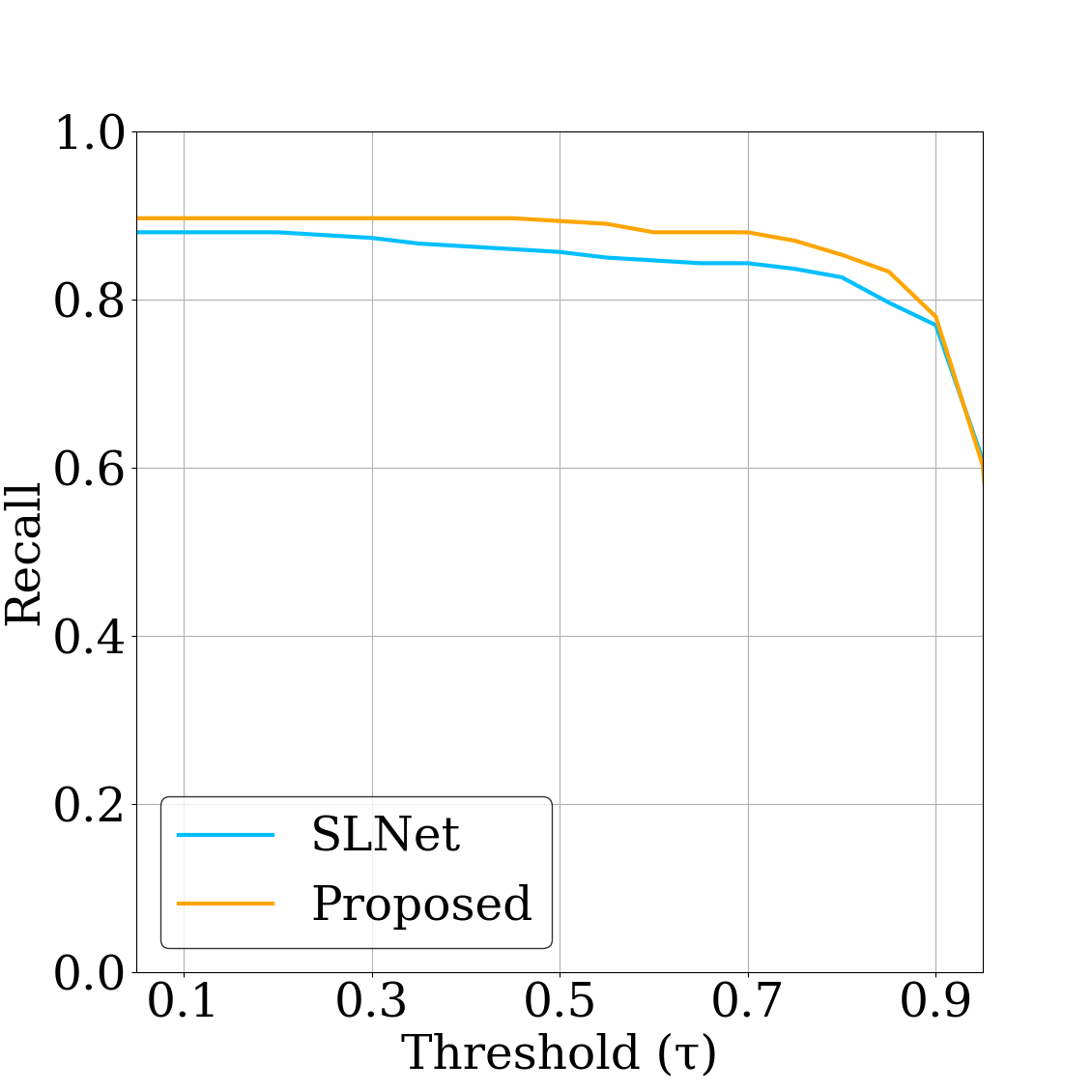}}
    \caption
    {
        Comparison of the accuracy, precision, and recall curves of the proposed DRM and the conventional SLNet in terms of the threshold $\tau$ on the SEL dataset.
    }
    \label{fig:auc_fig}
\end{figure*}

Fig.~\ref{fig:auc_fig} compares the accuracy, precision, and recall curves of the proposed DRM algorithm and the conventional SLNet algorithm~\cite{lee2017} on the SEL dataset. The proposed DRM outperforms SLNet in all three curves in the entire range of the threshold $\tau$. Table~\ref{table:comparison} reports the area under curve (AUC) performances of the accuracy, precision, and recall curves in Fig.~\ref{fig:auc_fig}, which are denoted by AUC$\_$A, AUC$\_$P, and AUC$\_$R, respectively. DRM provides higher AUC$\_$A, AUC$\_$P, and AUC$\_$R than SLNet by 2.54, 5.00, and 3.66, respectively. Table~\ref{table:comparison} also compares the performances on SEL\_Hard. For this comparison as well, we use the same DRM and SLNet, which are trained using the training images in the SEL dataset. Since SEL\_Hard consists of more challenging images, the performances are lower than those on SEL. Nevertheless, on SEL\_Hard, DRM outperforms SLNet by significant margins 7.09, 12.97, and 7.01 in terms of AUC\_A, AUC\_P, and AUC\_R, respectively.

\begin{table}[t]\centering

    \caption
    {
        Comparison of the AUC scores (\%) on the SEL and SEL\_Hard datasets.
    }
    \begin{tabular}[t]{+L{3.2cm}^C{1.2cm}^C{1.2cm}^C{1.2cm}^C{1.2cm}^C{1.2cm}^C{1.2cm}}
    \toprule
    \multirow{2}{*}{}  & \multicolumn{3}{c}{SEL} & \multicolumn{3}{c}{SEL\_Hard} \\
    \cmidrule(lr){2-4} \cmidrule(lr){5-7}
    & AUC\_A & AUC\_P & AUC\_R & AUC\_A & AUC\_P & AUC\_R \\
    \midrule
         SLNet           & 92.00 & 80.44 & 83.50 & 73.59 & 74.22 & 70.68\\
         Proposed DRM                      & 94.54 & 85.44 & 87.16 & 80.68 & 87.19 & 77.69 \\
    \bottomrule
    \end{tabular}
    \label{table:comparison}
\end{table}

\begin{figure*}[t]

    \begin{flushright}
    \subfloat {\raisebox{2em}{\rotatebox[origin=t]{90}{Ground-truth}}}\,\!
    \subfloat {\includegraphics[width=2.32cm,height=1.7cm]{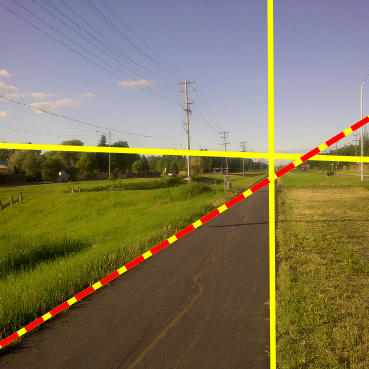}}\,\!\!
    \subfloat {\includegraphics[width=2.32cm,height=1.7cm]{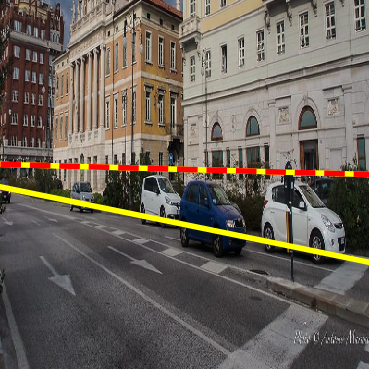}}\,\!\!
    \subfloat {\includegraphics[width=2.32cm,height=1.7cm]{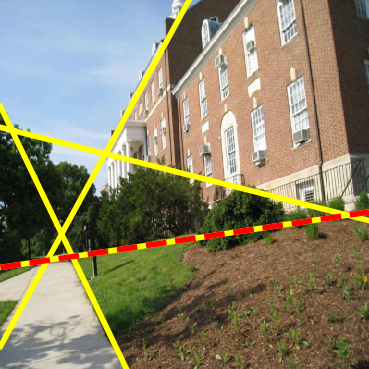}}\,\!\!
    \subfloat {\includegraphics[width=2.32cm,height=1.7cm]{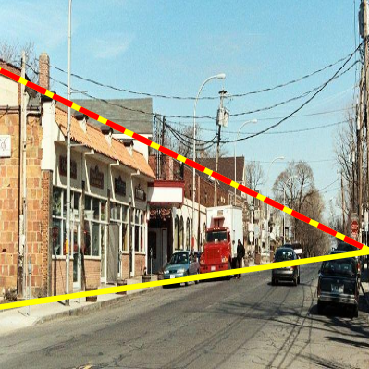}}\,\!\!
    \subfloat {\includegraphics[width=2.32cm,height=1.7cm]{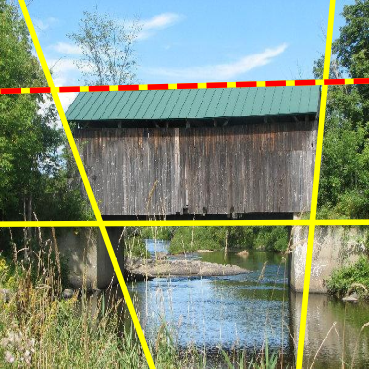}}\\[-2.9ex]

    \subfloat {\raisebox{2em}{\rotatebox[origin=t]{90}{SLNet}}}\,\!
    \subfloat {\includegraphics[width=2.32cm,height=1.7cm]{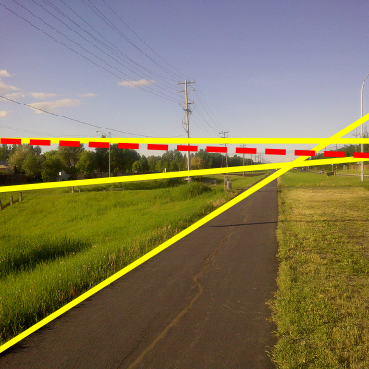}}\,\!\!
    \subfloat {\includegraphics[width=2.32cm,height=1.7cm]{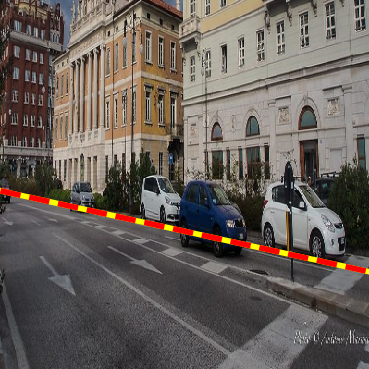}}\,\!\!
    \subfloat {\includegraphics[width=2.32cm,height=1.7cm]{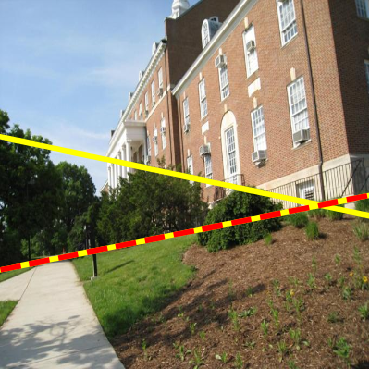}}\,\!\!
    \subfloat {\includegraphics[width=2.32cm,height=1.7cm]{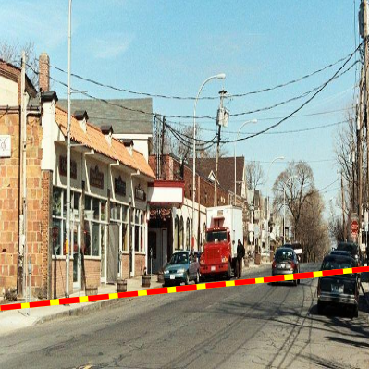}}\,\!\!
    \subfloat {\includegraphics[width=2.32cm,height=1.7cm]{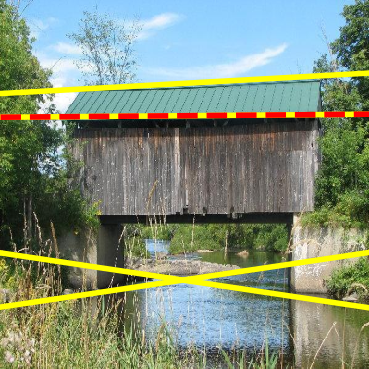}}\\[-2.2ex]

    \subfloat {\raisebox{2em}{\rotatebox[origin=t]{90}{Proposed}}}\!
    \subfloat {\includegraphics[width=2.32cm,height=1.7cm]{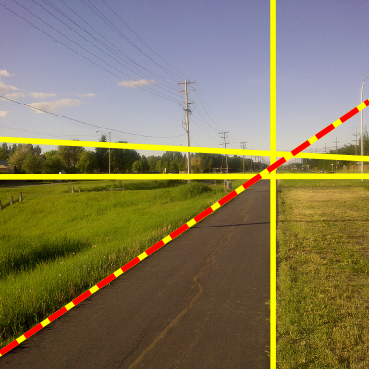}}\,\!\!
    \subfloat {\includegraphics[width=2.32cm,height=1.7cm]{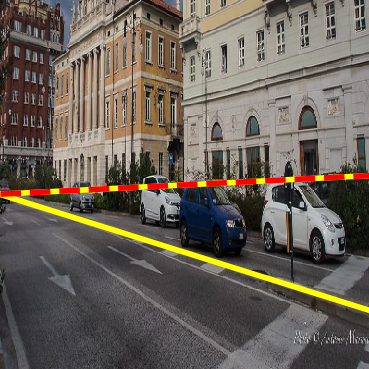}}\,\!\!
    \subfloat {\includegraphics[width=2.32cm,height=1.7cm]{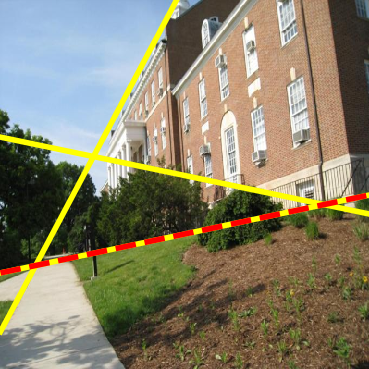}}\,\!\!
    \subfloat {\includegraphics[width=2.32cm,height=1.7cm]{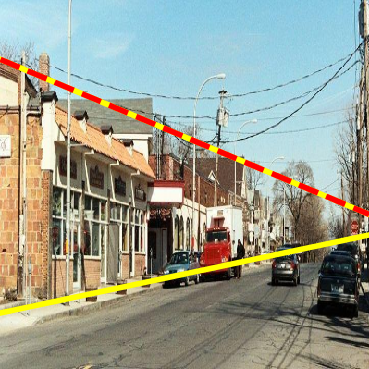}}\,\!\!
    \subfloat {\includegraphics[width=2.32cm,height=1.7cm]{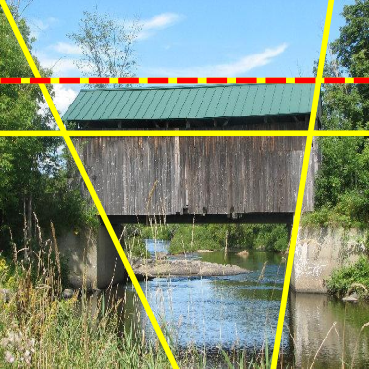}}

    \caption
    {
        Comparison of semantic line detection results. The left two images are from SEL, and the other three from SEL\_Hard. Primary and multiple semantic lines are depicted in dashed red and solid yellow, respectively.
    }
    \label{fig:result_fig}
    \end{flushright}
\end{figure*}

Fig.~\ref{fig:result_fig} compares detection results qualitatively. Compared to SLNet, the proposed DRM detects implied, as well as obvious, semantic lines more precisely. Also, DRM suppresses redundant lines more effectively. More detection results are available in the supplemental document.

\begin{table}[t]\centering

    \caption
    {
        The ablation studies in terms of the mirror attention module and comparative ranking and matching. AUC scores (\%) of primary and multiple semantic line detection are compared on the SEL dataset.
    }

    \begin{tabular}{+C{0.5cm}^L{8.0cm}^C{1.1cm}^C{1.1cm}^C{1.1cm}}
    \toprule
     & & AUC\_A & AUC\_P & AUC\_R \\
    \midrule
     \RomNum{1}.  & D-Net{(without attention)}+NMS & 92.48 & 83.37 & 84.76 \\
     \RomNum{2}.  & D-Net{(with attention, no flipped feature map)}+NMS & 93.24 & 81.33 & 84.86\\
     \RomNum{3}.  & D-Net{(with spatial-channel attention)}+NMS & 92.85 & 81.74 & 85.84 \\
     \RomNum{4}.  & D-Net{(with mirror attention)}+NMS & 93.38 & 83.98 & 86.23\\
     \RomNum{5}.  & D-Net{(with mirror attention)}+R-Net+M-Net & 94.54 & 85.44 & 87.16\\
    \bottomrule
    \end{tabular}
    \label{table:ablation}
\end{table}

\subsection{Ablation studies}
We conduct ablation studies to analyze the efficacy of the proposed D-Net, R-Net, and M-Net on the SEL dataset.

\subsubsection{Efficacy of mirror attention model:} Table~\ref{table:ablation} compares the performances of several ablated methods. First, to demonstrate the impacts of the mirror attention module in D-Net, we do not use the comparative ranking and matching (R-Net and M-Net). Instead, we adopt the non-maximum suppression (NMS) scheme in~\cite{lee2017}, which removes overlapped semantic lines based on low-level edge features. Method \RomNum{1} uses no attention module. Method \RomNum{2} uses the attention module in Fig.~\ref{fig:mirror_attention} but without concatenating a flipped feature map. Method \RomNum{3} replaces the mirror attention module with the spatial-channel attention in~\cite{woo2018}. As compared with no attention in \RomNum{1}, the two attention schemes in \RomNum{2} and \RomNum{3} improve the accuracy and recall scores but lower the precision score. On the contrary, the proposed mirror attention model in \RomNum{4} improves all three scores and also outperforms the two alternative schemes in \RomNum{2} and \RomNum{3}. Also, by comparing \RomNum{4} with \RomNum{2}, we see that the mirroring of feature maps across semantic lines is effective for emphasizing informative regions. More specifically, the mirroring improves AUC\_A, AUC\_P, and AUC\_R by 0.14, 2.65, and 1.37, respectively.

\subsubsection{Efficacy of R-Net and M-Net:}
By comparing methods \RomNum{4} and \RomNum{5}, we see that the proposed DRM algorithm provides $1.16$, $1.46$ and $0.93$ higher AUC\_A, AUC\_P, and AUC\_R scores, by employing R-Net and M-Net instead of NMS. This indicates that the proposed comparative ranking and matching is an effective approach to select reliable semantic lines and remove redundant ones.

\section{Applications}
We apply the proposed DRM algorithm to detect two kinds of semantically important lines: dominant parallel lines and reflection symmetry axes.

\begin{figure*}[t]

    \begin{flushright}

    \subfloat {\raisebox{2em}{\rotatebox[origin=t]{90}{SLNet}}}\,\!
    \subfloat {\includegraphics[width=2.32cm,height=1.7cm]{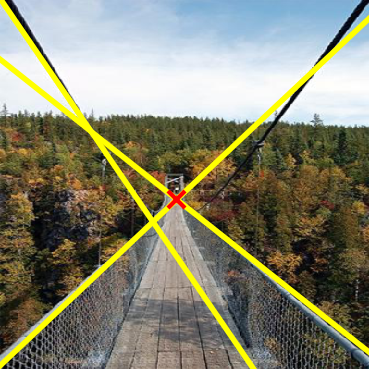}}\,\!\!
    \subfloat {\includegraphics[width=2.32cm,height=1.7cm]{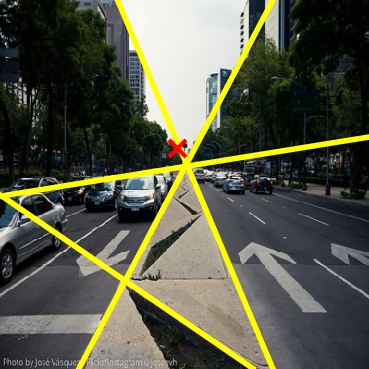}}\,\!\!
    \subfloat {\includegraphics[width=2.32cm,height=1.7cm]{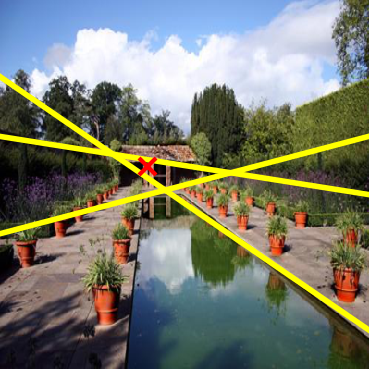}}\,\!\!
    \subfloat {\includegraphics[width=2.32cm,height=1.7cm]{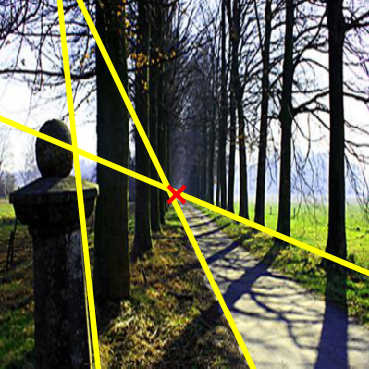}}\,\!\!
    \subfloat {\includegraphics[width=2.32cm,height=1.7cm]{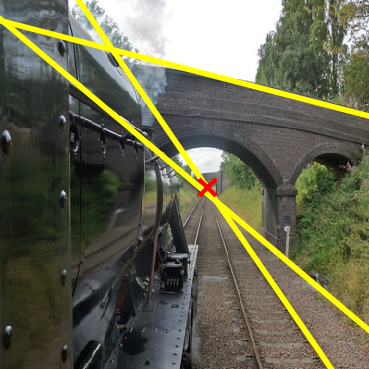}}\\[-2.1ex]

    \subfloat {\raisebox{2em}{\rotatebox[origin=t]{90}{D-Net+NMS}}}\!
    \subfloat {\includegraphics[width=2.32cm,height=1.7cm]{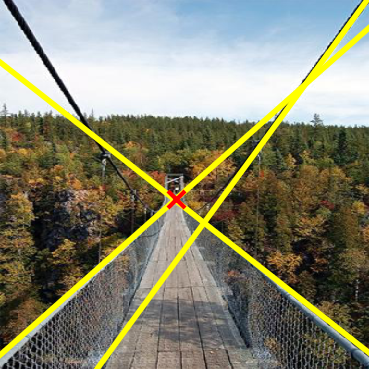}}\,\!\!
    \subfloat {\includegraphics[width=2.32cm,height=1.7cm]{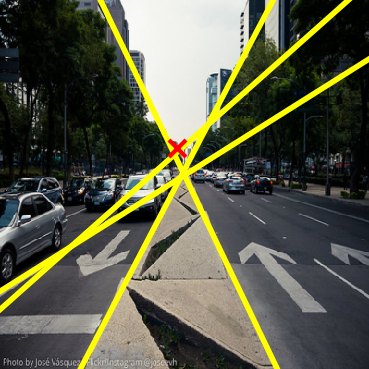}}\,\!\!
    \subfloat {\includegraphics[width=2.32cm,height=1.7cm]{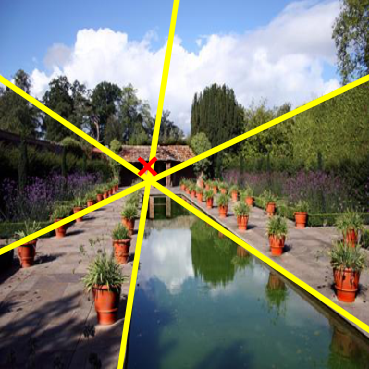}}\,\!\!
    \subfloat {\includegraphics[width=2.32cm,height=1.7cm]{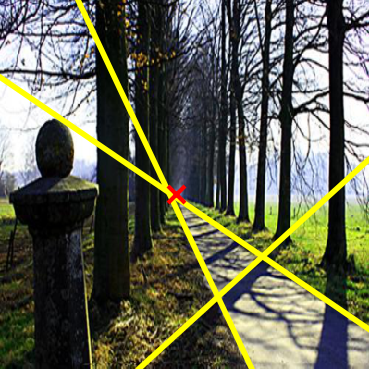}}\,\!\!
    \subfloat {\includegraphics[width=2.32cm,height=1.7cm]{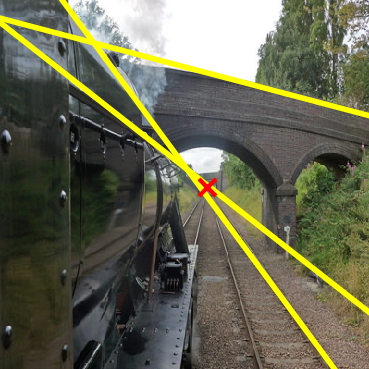}}\\[-2.5ex]

    \subfloat {\raisebox{2em}{\rotatebox[origin=t]{90}{Proposed}}}\!
    \subfloat {\includegraphics[width=2.32cm,height=1.7cm]{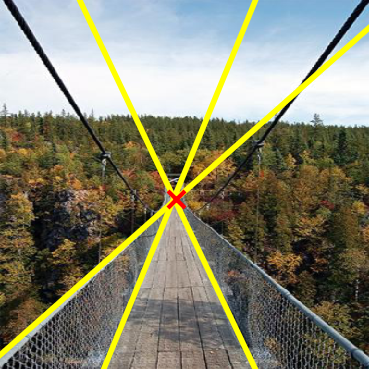}}\,\!\!
    \subfloat {\includegraphics[width=2.32cm,height=1.7cm]{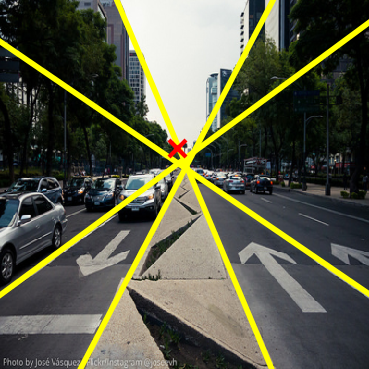}}\,\!\!
    \subfloat {\includegraphics[width=2.32cm,height=1.7cm]{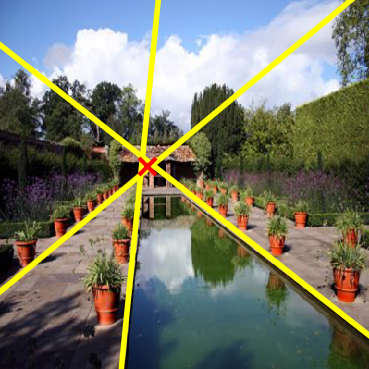}}\,\!\!
    \subfloat {\includegraphics[width=2.32cm,height=1.7cm]{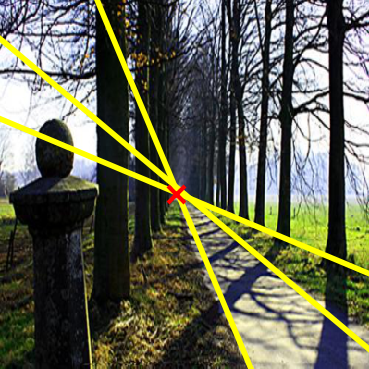}}\,\!\!
    \subfloat {\includegraphics[width=2.32cm,height=1.7cm]{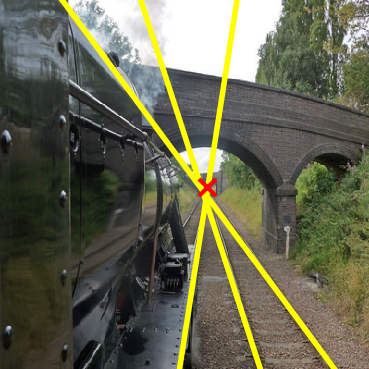}}

    \caption
    {
        Detection results of dominant parallel lines. For reference, the ground-truth vanishing points are depicted by red cross symbols.
    }
    \label{fig:Qualitative Comparison VP Line}
    \end{flushright}
\end{figure*}

\subsection{Dominant parallel lines}

When projected onto a 2D image, dominant parallel lines in the 3D world convey depth impressions, intersecting at a vanishing point (VP)~\cite{zhou2017}. Despite researches on the VP detection~\cite{zhou2017,zhou2019_nips}, a single VP is less informative for conveying the depth information than those projected parallel lines are. Hence, we apply the proposed DRM algorithm to detect dominant parallel lines in a 2D image. We first extract semantic lines using D-Net. To this end, D-Net is trained to detect semantic lines passing through VPs. Next, using R-Net, the primary semantic line is selected. In this application, M-Net is trained to declare a pair of lines, which are parallel in the 3D space, as `matched.' Then, we repeatedly select the semantic line yielding the highest `matched' score to the primary semantic line. To avoid overlapping, we remove the semantic lines whose mIoUs with an already selected one are greater than 0.85 after each selection.

\begin{table}[t]\centering
    \caption
    {
         AUC\_A scores (\%) in the dominant parallel line detection, according to the number $K$ of detected lines.
    }
    \begin{tabular}[t]{+L{2.5cm}^C{1.2cm}^C{1.2cm}^C{1.2cm}^C{1.2cm}^C{1.2cm}}
    \toprule
    $K$ & 1 & 2 & 3 & 4 & 5 \\
    \midrule
        SLNet~\cite{lee2017}    & 46.83 & 41.73 & 38.58 & 37.22 & 36.44 \\
        D-Net+NMS   & 52.62 & 48.93 & 46.24 & 44.71 & 43.30 \\
        Proposed    & 56.31 & 54.36 & 52.42 & 51.17 & 50.72 \\
    \bottomrule
    \end{tabular}

    \label{table:exp_vpline}
\end{table}

We assess the proposed algorithm on the AVA landscape dataset~\cite{zhou2017}. It contains 2,275 training and 275 test landscape images. For each image, a dominant VP and two dominant parallel lines are annotated. We declare a detected line as correct, when its distance to the ground-truth VP is smaller than a threshold. Then, we compute AUC\_A scores by varying the threshold. Table~\ref{table:exp_vpline} compares the AUC\_A scores of the conventional SLNet~\cite{lee2017}, `D-Net+NMS,' and the proposed algorithm according to the number $K$ of detected lines in each image. As more lines are selected in an image, the accuracy score is lowered. However, for every $K$, D-Net+NMS outperforms SLNet, which indicates that D-Net detects dominant parallel lines more precisely. Moreover, by employing R-Net and M-Net, the proposed algorithm further improves the performances. Fig.~\ref{fig:Qualitative Comparison VP Line} shows that the proposed algorithm detects dominant lines that pass through VPs accurately.

\begin{table}[t]\centering

    \caption
    {
         Comparison of AA scores (\%) for the dominant VP detection.
    }
   \begin{tabular}[t]{+L{3.0cm}^C{1.5cm}^C{1.5cm}^C{1.5cm}}
    \toprule
                       & AA1$^\circ$   & AA2$^\circ$   & AA10$^\circ$ \\
    \midrule
        Zhou \etal~\cite{zhou2017}     & 18.5 & 33.0 & 60.0 \\
        NeurVPS~\cite{zhou2019_nips}     & \bf{19.6} & \bf{35.9} & 65.9\\
        Proposed          & 8.6  & 22.9 & \bf{68.3} \\
    \bottomrule
    \end{tabular}
    \label{table:exp_vp}
\end{table}

Next, we detect a VP as the intersecting point of the first two selected lines. Table~\ref{table:exp_vp} compares this VP detection scheme with the existing methods\cite{zhou2017,zhou2019_nips}. Angle accuracies AA1$^\circ$, AA2$^\circ$, AA10$^\circ$ are used as the performance metrics, as done in \cite{zhou2019_nips}. Two performances of NeurVPS are reported in \cite{zhou2019_nips}. Table~\ref{table:exp_vp} includes their accuracies when the same training data as the proposed algorithm are used. Note that the proposed algorithm focuses on the detection of dominant lines and provides VPs as side results. In contrast, the existing methods are tailored for the VP detection. Therefore, when the tolerance angles are small (1$^\circ$ or 2$^\circ$), the proposed algorithm yields poorer accuracies than the existing methods. However, when the tolerance angle is 10$^\circ$, the proposed algorithm outperforms them. This indicates that the proposed algorithm can detect rough locations of VPs with a high recall rate, although it lacks the precision of the existing methods.

\subsection{Reflection symmetry axes}
Reflection symmetry is a common, but important visual property in various scenes, such as landscapes and man-made structures~\cite{liu2010}. However, since reflection symmetry axes are often highly implied or even invisible, their detection should exploit semantic regions around the axes. Accordingly, we train the proposed algorithm to detect the reflection symmetry axis of an image as the primary semantic line. More specifically, we train D-Net to extract symmetry axes as semantic lines, and R-Net to prioritize those axes among the detected lines. We empirically find that the Gaussian weighting in~(\ref{eq:filter}) and the residual attention in~(\ref{eq:residual_attention}) are less effective in this task. Hence, we exclude those operations from the mirror attention module in Fig.~\ref{fig:mirror_attention}. Then, we extract semantic lines using D-Net, and choose the most reliable one as the symmetry axis via \eqref{eq:top_rank} using R-Net.

\begin{figure*}[t]

    \centering
    \subfloat {\includegraphics[width=2.38cm,height=1.7cm]{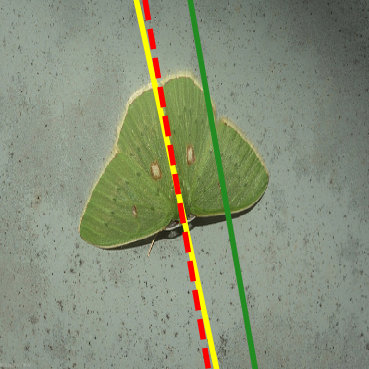}}\,\!\!
    \subfloat {\includegraphics[width=2.38cm,height=1.7cm]{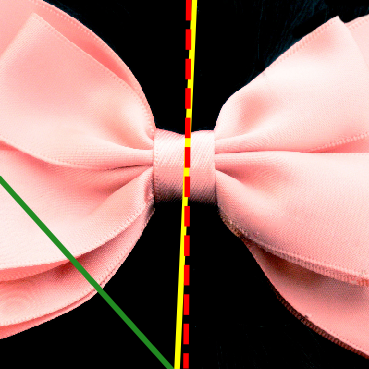}}\,\!\!
    \subfloat {\includegraphics[width=2.38cm,height=1.7cm]{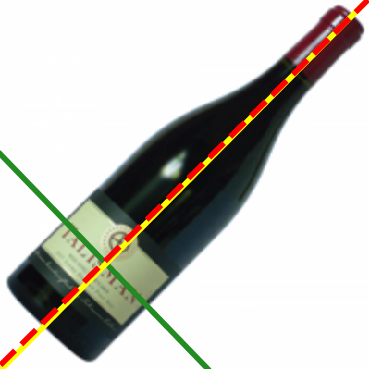}}\,\!\!
    \subfloat {\includegraphics[width=2.38cm,height=1.7cm]{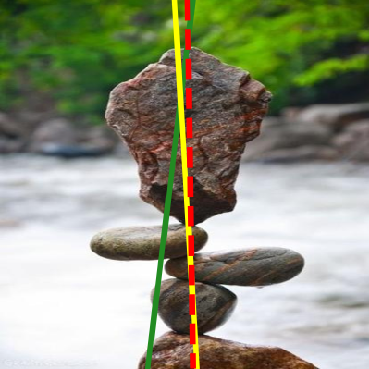}}\,\!\!
    \subfloat {\includegraphics[width=2.38cm,height=1.7cm]{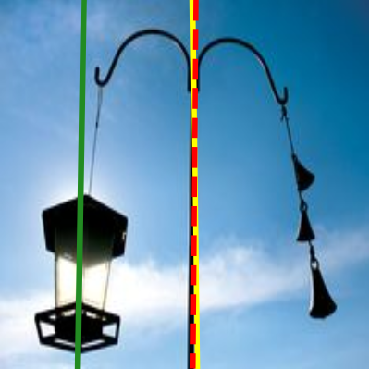}}

    \caption
    {
        Detection results of symmetry axes. The ground-truth axes are in red, the detection results of Loy and Eklundh~\cite{loy2006} are in green, and those of the proposed algorithm are in yellow.
    }
    \label{fig:asymm}

\end{figure*}

We test the proposed algorithm on three datasets: ICCV~\cite{funk2017}, NYU~\cite{cicconet2017}, and SYM\_Hard. ICCV provides 100 training and 96 test images, and NYU contains 176 test images. In SYM\_Hard, we collect 45 images from photo sharing websites~\cite{Flickr,Google}, each of which includes a reflection symmetry axis. The axis is implied, and the neighboring regions are not exactly symmetric. Thus, its detection is challenging. Since the proposed algorithm detects symmetry axes as primary lines, we compare the proposed algorithm with the existing methods~\cite{loy2006,cicconet2017_nyu,elawady2017,cicconet2017} using the AUC\_A metric. We train the proposed algorithm using the ICCV training images and use it to assess the performances on all three datasets.  Table~\ref{table:exp_sym} compares the results. On ICCV, NYU, and SYM\_Hard, the proposed algorithm outperforms the existing methods by at least 0.83, 1.93, and 2.74, respectively. Fig.~\ref{fig:asymm} compares detection results of the proposed algorithm with those of Loy and Eklundh~\cite{loy2006}. The proposed algorithm detects symmetry axes more robustly.  More experimental results are available in the supplemental document.

\begin{table}[t]\centering
    \caption
    {
        Comparison of AUC\_A scores (\%) of the symmetry axis detection.
    }
    \begin{tabular}[t]{+L{3.4cm}^C{1.5cm}^C{1.5cm}^C{1.9cm}}
    \toprule
                    & ICCV    & NYU & SYM\_Hard \\
    \midrule
        Cicconet \etal~\cite{cicconet2017_nyu}         & 80.80      & 82.85    & 68.99\\
        Elawady \etal~\cite{elawady2017}              & 87.24      & 83.83    & 73.90\\
        Cicconet \etal~\cite{cicconet2017}             & 87.38      & 87.64    & 81.04\\
        Loy \& Eklundh \cite{loy2006}                  & 89.77      & 90.85    & 81.99\\
        Proposed                      & \bf{90.60} & \bf{92.78} & \bf{84.73}\\

    \bottomrule
    \end{tabular}
    \label{table:exp_sym}
\end{table}

\section{Conclusions}
We proposed a novel semantic line detector using D-Net, R-Net, and M-Net. First, D-Net extracts semantic lines using the mirror attention module. Second, R-Net selects the most semantic line through ranking. Third, M-Net removes redundant lines overlapping with the selected one through matching. The second and third steps are alternately performed to yield reliable semantic lines as output. Experimental results demonstrated that the proposed DRM algorithm outperforms the conventional SLNet significantly. Moreover, it was shown that the proposed algorithm can be applied to successfully detect two important kinds of semantic lines:  dominant parallel lines and reflection symmetry axes.

\section*{Acknowledgements}

This work was supported in part by the Agency for Defense Development (ADD) and Defense Acquisition Program Administration (DAPA) of Korea under grant UC160016FD and in part by the National Research Foundation of Korea (NRF) through the Korea Government (MSIP) under grant NRF-2018R1A2B3003896.

\bibliographystyle{splncs04}
\bibliography{3397}
\end{document}